\pdfoutput=1
\documentclass[11pt]{article}

\usepackage[]{acl}

\usepackage{times}
\usepackage{latexsym}

\usepackage[T1]{fontenc}
\usepackage[utf8]{inputenc}

\usepackage{microtype}

\usepackage{inconsolata}

\usepackage{colortbl}
\usepackage{booktabs}
\usepackage{multirow}
\usepackage{caption}
\usepackage{subcaption}
\usepackage{enumitem}
\usepackage{amssymb,amsmath,bm}
\DeclareMathOperator*{\argmax}{argmax}

\definecolor{Gray}{gray}{0.9}
\usepackage{bbding}
\usepackage{amssymb}
\usepackage{graphicx}
\usepackage{xspace}
\usepackage[normalem]{ulem}
\usepackage{verbatim}

\title{Fast Streaming Transducer ASR Prototyping via\\ Knowledge Distillation with Whisper}

\author{{\bf Iuliia Thorbecke\footnotemark[1]$^{1,2}$}
{\bf Juan Zuluaga-Gomez\footnotemark[1]$^{1,3}$}
{\bf Esaú Villatoro-Tello$^{1}$} \\
{\bf Shashi Kumar$^{1,3}$}
{\bf Pradeep Rangappa$^{1}$}
{\bf Sergio Burdisso$^{1}$} \\
{\bf Petr Motlicek$^{1,4}$}
{\bf Karthik Pandia$^{5}$}
{\bf Aravind Ganapathiraju$^{5}$}
\\
$^{1}$ Idiap Research Institute, Switzerland;
$^{2}$ University of Zurich, Switzerland; \\
$^{3}$ EPFL, Switzerland;
$^{4}$ Brno University of Technology, Czech Republic;
$^{5}$ Uniphore, India \\
\normalsize\texttt{iuliia.nigmatulina@idiap.ch}
\normalsize\texttt{juan.zuluaga@eu4m.eu}
}

\begin{document}
\maketitle

\renewcommand*{\thefootnote}{\fnsymbol{footnote}}
\footnotetext[1]{Equal contribution. Order is determined by a coin flip.}
\renewcommand*{\thefootnote}{\arabic{footnote}}

\begin{abstract}
The training of automatic speech recognition (ASR) with little to no supervised data remains an open question.
%In some cases, this involves a pre-train and then fine-tune stage that requires large data and computational budget.
In this work, we demonstrate that streaming Transformer-Transducer (TT) models can be trained from scratch in consumer and accessible GPUs in their entirety with pseudo-labeled (PL) speech from foundational speech models (FSM).
This allows training a robust ASR model just in one stage and does not require large data and computational budget compared to the two-step scenario with pre-training and fine-tuning. We perform a comprehensive ablation on different aspects of PL-based streaming TT models such as the impact of (1)~shallow fusion of n-gram LMs, (2)~contextual biasing with named entities, (3)~chunk-wise decoding for low-latency streaming applications, and (4)~TT overall performance as the function of the FSM size. Our results demonstrate that TT can be trained from scratch without supervised data, even with very noisy PLs. We validate the proposed framework on 6 languages from CommonVoice and propose multiple heuristics to filter out hallucinated PLs.
\end{abstract}

\section{Introduction}
\label{sec:introduction}

%\TODO{shou;d we change the title and a bit the paper towards something like offering a "framework" for developing streaming solutions, including contextual biasing, etc. }

There are many challenges when developing automatic speech recognition (ASR) engines for industrial applications, including (1)~large-scale databases that generalize across multiple domains; (2)~inference under challenging low-latency settings; and (3)~lightweight ASR model size to minimize deployment costs. 
While the first has been solved by training large acoustic foundational speech models (FSM) with massive databases \citep{conneau21_xlsr_53,pratap2023_mms}, the latter two strongly relate to architectural choices, e.g., using Connectionist Temporal Classification (CTC) \citep{graves2006connectionist_ctc} or transducer-based \citep{graves2012sequence_rnn_t} modeling. 
%16.675 14.975
\begin{figure}[t]
    \centering
\includegraphics[width=0.99\linewidth]{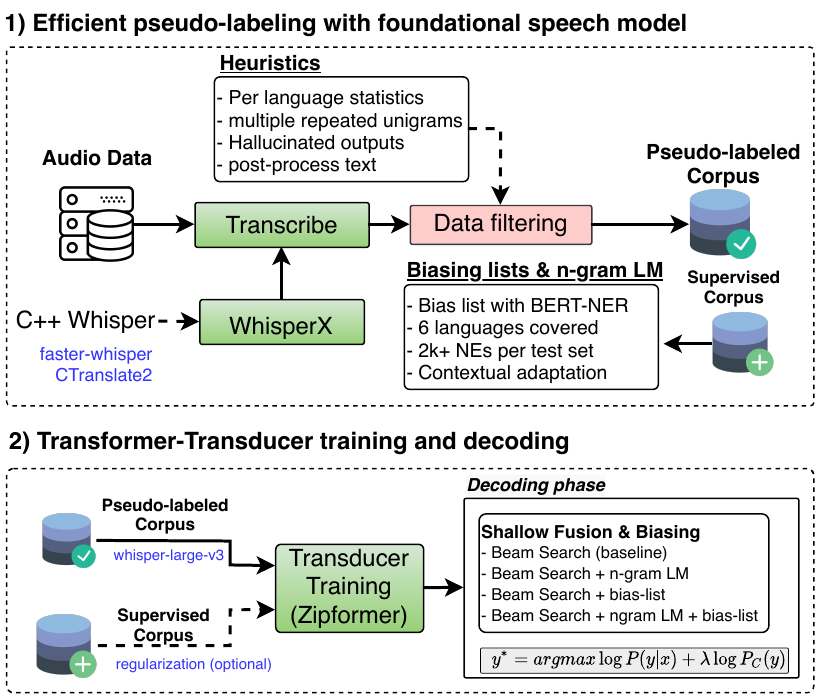}
    \caption{Proposed framework for efficient and fast streaming ASR prototyping with pseudo-labeled data. Transducer models are further improved via shallow fusion of n-gram LMs and contextual biasing of target named entities.}
    \label{fig:approach}
\end{figure}

In industrial applications, large supervised databases in target domains are not always available, thus several techniques have been proposed to develop robust ASR models with small supervised corpora:
(1)~data augmentation~\citep{park2019specaugment,bartelds-etal-2023-making};
(2)~only-audio self-supervised pre-training with large databases and fine-tuning with small corpora~\citep{baevski2020wav2vec2,conneau21_xlsr_53,zuluaga2023does};
(3)~pseudo-label then fine-tune, e.g., semi-supervised learning~\citep{zhu2023alternative,lugosch2022pseudo,zuluagagomez21_interspeech} and weakly supervised learning ~\citep{whisper}. Most of the approaches target the attention-based encoder-decoder (AED)~\citep{watanabe2017hybrid} or CTC models. 
% We focus on the last approach, as it is more straightforward and do not need much tuning and do not need large computation clusters. Also, because models can be trained from scratch and can work for very specific domains. 
Even though these two architectures have shown impressive results on multiple benchmarks (e.g., Whisper~\citep{whisper}), they still lag in streaming settings~\citep{prabhavalkar2023end}. 

The Transformer-Transducer architecture~\citep{yeh2019_tt} is widely exploited for industrial uses that require streaming decoding because the transducer decoder naturally supports
streaming~\citep{cascaded-encoder-streaming,rnnt-las-rescoring}. However, the transducer used to be harder to train compared to AED and CTC, thus, it was less explored in the community, until it was shown to achieve a performance as close as AED models~\citep{sainath2020streaming_on_device}. The transducer models consist of an encoder, predictor and joint networks. Using a Transformer \citep{vaswani2017attention} encoder leads to a Transformer-Transducer (TT) architecture~\citep{battenberg2017exploring_rnnt,yeh2019_tt,zhang2020_tt}. When trained from scratch, the TT models require sufficient amounts of supervised datasets in the target language and domain~\citep{fastconformer-streaming,cascaded-encoder-streaming}. At the same time, fine-tuning a large pre-trained model, even when using a transducer decoder, would not allow streaming decoding.

% In this work, we target two questions partly unanswered by the research community: (1) Could we prototype a streaming Transformer-Transducer ASR models in one GPU day? (2) Can we train TT models with pure pseudo-labeled (PL) data? 
In this work, we focus on two questions partly unanswered by the research community: (1) Could we quickly prototype a streaming TT model on a single accessible GPU? (2) Can we train TT models with only pseudo-labeled (PL) data? We target the streaming scenario, which is by nature more challenging than standard offline (full attention) decoding~\citep{sainath2020streaming_on_device}. Despite the robustness of AED models in the offline scenario, they still require a large amount of supervised data. 
%Prior work aims to bring AED to streaming setups~\citep{kumar24_xlsr_transducer}. 
Here, we use TT models~\citep{yeh2019_tt}, where the challenge arises on the fact that these do not include a self-supervised stage,\footnote{\citet{chiu2022self_best_rq} explore to warm start the encoder with a pre-trained SSL-based model, albeit closed source model.} i.e., needing audio-text pairs always. We demonstrate that TT models can be trained entirely from scratch with PLs generated from Whisper~\citep{whisper} while attaining competitive performance in streaming scenarios. The overall proposed approach is illustrated in Figure~\ref{fig:approach}.

\paragraph{Contributions:}
\begin{itemize}[nosep,topsep=-\parskip]
    \item We propose a framework for full-stack rapid development of ASR streaming solutions from scratch with low-to-zero supervised resources;
    \item comprehensive study of TT performance as a function of the pseudo-labels quality, for both, online and offline settings;
    \item robust heuristics to filter out noisy and hallucinated PLs from FSM;
    \item evaluation of the impact of shallow fusion with external n-gram LM and contextual biasing for named entities;
    % \item ablation studies, including mixing of the small supervised subset as regularization during training;
    \item experimentation and validation on 6 languages from CommonVoice.
\end{itemize}

\section{Related Work}
\label{sec:related-work}

Developing robust ASR systems for low-latency online settings with little to no supervised data is still an open challenge in the community. In this section, we introduce the most prominent approaches to overcome these issues. 

\paragraph{From Encoder-Decoder to Transducer models} \quad
One of the key advantages of transducer models over encoder-decoder relies on the fact that it supports streaming decoding. Not until recently, it has been demonstrated that these models can surpass standard AED systems ~\citep{sainath2020streaming_on_device}. There have been multiple breakthroughs that have made transducer training easier, such as (1)~pruned transducer loss ~\citep{kuang22_pruned_rnn_t}, (2)~better architectures, e.g., FastConformer \citep{fastconformer};
%and HyperConformer \citep{mai23hyperconformer}
and (3)~from the modeling side, e.g., model pruning, sparsification~\citep{yang2022omni_pruning}, and quantization \citep{sainath2020streaming_on_device}. However, little to no work has been done on fast TT model prototyping (few GPU-days) with pure pseudo-labeled data. 

\paragraph{Pseudo-labeling in ASR} \quad Semi-supervised learning \citep{zhang2020pushing,park2020improved,higuchi21_interspeech}, pseudo labeling \citep{zavaliagkos1998untranscribed_asr,likhomanenko2020slimipl,hwang2022pseudo}, and weakly supervised learning \citep{whisper} are a family of methods aiming to partly alleviate the burden of lack of labeled data for supervised ASR training. These methods have shown promising word error rate (WER) improvement in multiple settings and languages. In practice, a teacher model is trained on an audio-text paired corpus $D_l=\{X_{i},Y_{i}\}$. Then, it is used to pseudo label a much larger unlabeled only audio corpus, $D_{pl}=\{X_{i},Y^{*}_{i}\}$. Afterward, usually a smaller model~\citep{barrault2023seamlessm4t} can use $D_l$ and $D_{pl}$ for supervised training or fine-tuning~\citep{hsu2021hubert}.

The main difference between most of the previous studies and our approach proposed in the paper is that typically \textit{iterative training} is used \cite{zhang2020pushing,park2020improved,hwang2022pseudo}. A multi-stage strategy combining self- and semi-supervised learning eventually results in strong pseudo-labels. In the present paper, we focus on the performance that can be achieved ``out-of-the-box'' by using already available FSM and training transducer models from scratch and only once. Thus, this approach allows for minimizing the overall computational cost, i.e., one-stage training and applying improvement methods, such as decoding with shallow fusion. We aim to reveal the potential of the FSM in pseudo-label generation and demonstrate what performance can be reached with minimal training efforts.

PLs, however, are often noisy and bounded by the quality of a teacher model, whereas their use might result in suboptimal final performance in the models. This can be solved by either filtering out the nosiest samples or increasing the teacher model size to improve their quality.\footnote{We assume that a larger model, trained under the same conditions and with increased data, will attain lower WERs.} Several approaches to improve the PL quality include improving loss functions \citep{zhu2023alternative,gao2023bypass},
pairing online and offline models at training time \citep{higuchi21_interspeech}, and continuous single-language ~\citep{likhomanenko2022continuous,berrebbi2022continuous} and multilingual pseudo-labeling setting~\citep{lugosch2022pseudo}.

% \paragraph{Streaming Decoding Challenges} We target the most challenging scenario, which is transducer training for deployment on streaming scenarios, where we are bounded by challenges such as real-time decoding with low latency and training with few or no supervised data. \TODO{to fill in, or not?}

\paragraph{Knowledge Distillation with Large Models} \quad Knowledge distillation (KD), or teacher-student training~\citep{watanabe2017student}, is a very well-known technique to distill knowledge from a large model into a smaller model~\citep{hinton2015distilling}. The former is considered the \textit{Teacher} and the latter is the \textit{Student}. In this framework, we first train the teacher model with the correct label (e.g., supervised training)~\citep{takashima2018investigation} or in a self-supervised manner. The student model is then trained with the posterior distributions of the pre-trained teacher model~\citep{chebotar16_interspeech}. There has been prior work on KD for CTC~\citep{takashima2018investigation} and AED models with Whisper~\citep{whisper}~\citep{gandhi2023distilwhisper,ferraz2024multilingual_distilwhisper} and Transducer models~\citep{panchapagesan2021efficient}. Similarly, work to distil offline transducer models into online has been explored by~\citet{kurata2020knowledge} or from self-supervised models~\citep{yang2022knowledge}.

In our work, we focus on sequence-level KD, which means we use the one-best hypothesis from the teacher model instead of using the posterior distribution. This approach has some benefits: (1)~no need to cache the teacher model or its outputs into memory; (2)~no need to modify the current ASR training pipelines; (3)~overall faster ASR training w.r.t teacher-student based KD, where we can leverage highly optimized inference pipelines--including model quantization--for PL generation, e.g., WhisperX~\citep{bain23_whisperX}. All this results in pseudo-labeling that meets the needs for fast prototyping for standard industrial applications.

\section{Experimental Setup}
\label{sec:experimental-setup}

This section describes the datasets, TT architecture, details for training with pseudo-labeled data, effective integration of language model and contextual biasing with shallow fusion, and metrics we use for evaluation. 

\subsection{Pseudo Labeling with Whisper}
\label{subsec:pseudo-labeling-with-whisper}

Our core contribution is the fast prototyping of TT streaming ASR trained exclusively on pseudo-labeled data. We select the Whisper model as our teacher model~\citep{whisper} due to its strong performance across multiple benchmarks. In addition, Whisper provides models at different parameter scales.

\paragraph{Decoding with WhisperX pipeline} \quad We use the WhisperX pipeline~\citep{bain23_whisperX} across all the experiments to generate PLs. It is composed of (1)~a voice activity detection step to segment long-form audio; (2)~batching multiple segments for efficient inference; (3)~model quantization of Whisper and C++ implementation on FasterWhisper\footnote{\url{https://github.com/SYSTRAN/faster-whisper}} which uses CTranslate2 for fast decoding;\footnote{\url{https://github.com/OpenNMT/CTranslate2/}} (4)~model inference and word level alignment. Note that we pseudo-label each training corpus with 5 Whisper model sizes, i.e., whisper-tiny, base, small, medium, and large-v3.

\paragraph{Data filtering heuristics} \quad
We developed multiple data selection heuristics ($H$) to filter out noisy and hallucinated PLs. \textbf{$H1$:} remove PL if composed of the same unigram three or more times. \textbf{$H2$:} compute maximum word length from supervised training corpus and remove utterances with one or more PLs larger than the max threshold.\footnote{See the per language proposed thresholds in appendix~\ref{sec:appendix-filtering-stage}.} \textbf{$H3$:} compute $word_{ratio}$\footnote{Number of words divided by utterance duration [seconds].} and filter out samples with $word_{ratio}$ less than 1 or more than 4. \textbf{$H4$:} verbalize all the numbers from the pseudo-labels, remove punctuation and normalize following the CommonVoice recipe in Lhotse~\citep{zelasko2021lhotse}. These heuristics are applied for every training corpora. Similar heuristics are proposed in~\citep{barrault2023seamlessm4t}.

\subsection{Transformer-Transducer Training}
\label{subsec:tt_training}

We train Transformer-Transducer models from scratch for each language and dataset. We use stateless predictor~\citep{stateless-predictor} and Zipformer encoder model~\citep{yao2023zipformer} with the latest Icefall Transducer recipe and its default training hyper-parameters.\footnote{\url{https://github.com/k2-fsa/icefall/tree/master/egs/librispeech/ASR/zipformer}.} This includes \textit{ScaledAdam} optimizer~\citep{kingma2014adam}, learning rate scheduler with a 500-step warmup phase~\citep{vaswani2017attention} followed by a decay phase (each 7.5k steps and 3.5 epochs), as in~\citet{yao2023zipformer}. The neural TT model is jointly optimized with an interpolation of simple and pruned RNN-T loss~\citep{kuang22_pruned_rnn_t,graves2012sequence_rnn_t} and CTC loss~\citep{graves2006connectionist_ctc} ($\lambda=0.1$), according to: 
\begin{equation}
    \label{eq:joint-rnnt-pruned-ctc-loss}
    \mathcal{L} =(1 - \lambda) \cdot \mathcal{L}_{RNN\-T} + \lambda  \cdot \mathcal{L}_{CTC}.
\end{equation}
We use an effective batch size of 600s with a gradient accumulation of 1, the peak learning rate is $lr=5.0e^{-2}$ and we train each TT for 30 epochs on a single RTX 3090 GPU with only PLs.\footnote{We also run an experiment valuable for the industrial domain. It includes a thorough analysis of PLs quality for the call-center domain, see Appendix~\ref{sec:appendix-call-center-use-case}.} Training takes between 1 and 2 days. During the decoding, we use a beam size of 4.

\paragraph{Regularization with supervised data} \quad We perform experiments where along with PLs we mix in 100h of randomly selected supervised data from the train set $D_l$ during training. We compute mixing weights between $D_l$ and $D_{pl}$ so each training batch contains at least one sample from $D_l$. This is achieved with \textit{CutSet.Mux} function from Lhotse~\citep{zelasko2021lhotse}.\footnote{It lazily loads two or more datasets and mixes them on the fly according to pre-defined mixing weights.} 
All the experiments that uses PL and supervised data are denoted with \textbf{\textit{+sup. [100h]}}, otherwise, the model is trained with PL only. As an ablation experiment, we also test the performance by scaling up supervised data to 200h and 400h when using the weakest FSM, i.e., whisper-tiny. This experiment aims to (1)~compensate for very low-quality PLs, and (2)~demonstrate that Whisper PLs (from the largest models) are of sufficient quality for transducer training without any supervised data. 

\paragraph{Enabling streaming decoding with multi-chunk training} All the models proposed in this work can perform streaming decoding. This is achieved by performing chunk-wise multi-chunk training. During training, we use causal masking of different sizes to enable streaming decoding under different low-latency configurations~\citep{swietojanski2023variable_attention,kumar24_xlsr_transducer}. Specifically, we 
rely on two lists: chunk-size=\{640ms,1280ms,2560ms,full\} and \mbox{left-context-frames=\{64,128,256,full\}}.\footnote{The effective number of left context chunks is computed as $left\_context\_frames // chunk\_size$.} At training time, we randomly select the chunk size and the left context chunks for each batch. This enables the final model to work on a wide variety of streaming settings. At test time, we select 13 different decoding configurations ranging from 320 ms\footnote{Decode chunk size of 320ms is more challenging as it has not been used during training.} to 2560 ms chunks (see App.~\ref{sec:appendix:streaming-configurations}). 

\subsection{Language Modeling and Contextual Biasing}

Leveraging more text data and context information with language model and keywords integration can considerably improve ASR performance. Since in our set-up, we assume that we have zero (or very little) supervised data, using extra unpaired text data would not contradict the original constraints. At the same time, relying mainly on pseudo-labels, we see text knowledge integration as an opportunity to make our models more reliable and robust. The widely used method of LM integration during the decoding is \textit{shallow fusion} (SF)~\citep{aleksic2015bringing,kannan2018analysis, zhao2019shallow,jung2022spell}. SF means log-linear interpolation of the score from the ASR model with an external separately optimized LM at each step of the beam search:
\begin{equation}
  y^{\ast} = \argmax \log P(y|x) + \lambda \log P_{LM}(y),
\end{equation}
where $P_{LM}(y)$ is an external LM and $\lambda$ is a hyperparameter to control the impact of the LM on the overall model score.

To gain more possible improvement from the text information, we explore three options with the SF: (1)~word-level n-gram LM, (2)~named-entities, (3)~combination of word-level n-gram LM and named-entities. We choose n-gram over neural network (NN) LMs, as the use of NN-LMs would be impractical in low-latency streaming scenarios due to the size of the models. Named entities are extracted automatically and considered as keywords forming biasing lists: for more details see Section~\ref{sec:cv_database}.

\paragraph{Shallow fusion with Aho-Corasick}
One of the drawbacks of LM fusion is that it typically slows down the decoding time during inference, especially when using bigger NN-LMs. Since we focus on streaming ASR models in this paper, any potential increase in inference time is critical for us.
Recent studies demonstrated that SF implemented with the Aho-Corasick (AC) algorithm \citep{aho1975efficient} is fast and optimized when used for the keyword biasing~\citep{guo2023improved}. Thus, we use the AC implementation from Icefall\footnote{\url{https://github.com/k2-fsa/icefall/blob/master/icefall/context_graph.py}} to integrate key named entities (NE) and word-level n-gram LMs during the decoding.
%Recent studies demonstrated that shallow fusion implemented with the Aho-Corasick (AC) algorithm \citep{aho1975efficient} and when using light n-gram LMs introduces almost no delay in inference compared to the baseline \citep{guo2023improved,our_paper}. Thus, we use the same implementation to integrate n-gram LM and key named entities (NE) during the decoding\citep{our_paper}.

The Transducer model we use outputs its hypotheses at the subword level and, in this case, an external LM is also typically trained on subwords. In our experiments, to benefit from the word-level statistics, we integrate word-based n-gram external LMs. Such integration from word to subword level is possible with the AC implementation. First, the LM n-grams are converted into strings of subword units with SentecePieces;\footnote{\url{https://github.com/google/sentencepiece}} second, the subword units are used to build an AC prefix trie including LM weights in the probability domain.

When a string match occurs between a model prediction and a string in the prefix trie, the log probability of the matching hypothesis is augmented by the LM weight. To obtain positive cost, we convert the logarithmic LM weights (e.g., from ARPA) back to probabilities by taking an exponent.
In the case of context biasing, SF works in the same way but instead of LM weights a fixed bias cost is added to each matched arc. Typically when applying context biasing alone, we set such cost to 0.7.\footnote{For contextual biasing with NEs, we tested the biasing costs = \{0.1, 0.3,0.5,0.7,1.0,1.5,2.0\}. 0.7 performed systematically better in all scenarios.}
%\TODO{I used a cost of 0.7 for all experiments that yielded the best results.}
For SF with combined n-gram LM and biasing list, we still use LM weights, bias cost, and a single prefix tree. Using a single prefix tree has the advantage of faster running time, which is relevant for streaming models. We tune the biasing cost on the dev sets and set it differently when a biased entity is present in the LM vs when it is not\footnote{For SF of n-gram LM combined with NEs, we tested the biasing following costs: inLM = \{0.5,1.0,1.5,2.0\};
notInLM= \{0.5,1.0,1.5,2.0\}. inLM=0.5 and notInLM=1.5 performed systematically better in all settings.}:
\[
C = 
\begin{cases}
  \alpha_{outLM} & \text{if NE is not in LM,} \\
  exp(LMw) + \alpha_{inLM} & \text{if NE is in LM,} \\
  exp(LMw) & \text{otherwise.}
\end{cases}
\]

\paragraph{Language modeling} \quad For LM SF, we train tri-gram word-level LMs with SRILM \citep{srilm}. To train n-gram LMs, we use text data from the corresponding train sets.
%For the low-resource languages that do not have enough text data to gain statistics to train LM, additional text data from multilingual librispeech~\citep{pratap20_multilingual_librispeech} was used. \TODO{more details and update}.
All the train texts are uppercased and normalized to contain only unicode characters. 

\noindent \textbf{Evaluation protocol} \quad For evaluation, we use the standard word error rate (WER) metric for ASR which is the lower the better.
%Additionally, we assess the accuracy and WER specifically for named entities (NEs), termed \textit{NE-A} and \textit{NE-WER}. The NE metrics are determined only on strings that contain NEs. NE accuracy is measured in a binary way: "yes" if the NE is entirely recognized correctly, and "no" if there is any error within the NE. 

\subsection{Databases}

Here, we introduce the datasets used for fast ASR prototyping and describe the process of generating biasing lists for each language.

\subsubsection{CommonVoice Database}
\label{sec:cv_database}
The CommonVoice dataset comprises several thousand hours of audio in more than 100 languages~\citep{ardila2019_commonvoice_corpus}. To the best of our knowledge,\footnote{According to the discussions in the official OpenAI-Whisper GitHub repository: \url{https://github.com/openai/whisper/discussions/349}, \url{https://github.com/openai/whisper/discussions/2305}.} the CommonVoice data was not used for training Whisper model and can be used for zero-shot evaluation. In our case, it is an important point, as using unseen data for generating PLs provides a more realistic estimation of the proposed approach performance.\footnote{Although officially the CommonVoice data is not included in the training data for Whisper, we realize the possibility of some CommonVoice data still being seen by the model through other sources and in a small amount.} For experimentation, we select six languages from CommonVoice-v11~\citep{ardila2019_commonvoice_corpus} \footnote{CommonVoice-v11: cv-corpus-11.0-2022-09-21} which have sufficient data for training ASR and language models:
Catalan (CA),
English (EN),
German (DE),
French (FR),
Spanish (ES), and Italian (IT). We use the official train sets and report WERs on the official test sets. See Table~\ref{tab:train_and_test_sets} for further statistics. 

\begin{table}[t]
    \caption{Train and test sets statistics with context information per CommonVoice language. $^{\dagger}$total of unique entities per test set after removing unigrams shorter than 5 characters. $^{\ddagger}$number of utterances in the test set with at least one named entity. 
    }
    \label{tab:train_and_test_sets}
    \centering
    \resizebox{1\linewidth}{!}{
    \begin{tabular}{l | c| ccc}
    \toprule
    Lang & Train set & \multicolumn{3}{|c}{Test set stats \& Named entities} \\
    \cmidrule(lr){2-2}
    \cmidrule(lr){3-5}
     (code) & \textbf{[hr]} & \textbf{utt/hr} & \textbf{unique}$^{\dagger}$ & \textbf{nb. utt}$^{\ddagger}$ \\
     \midrule
    EN & 1000 & 16K/27 & 6921 & 6442 \\
    CA & 1200 & 16.3K/28 & 2108 & 2607 \\
    FR & 600 & 16K/26 & 6035 & 7486 \\
    DE & 600 & 16K/27 & 6949 & 8491 \\
    ES & 317 & 15.5K/26 & 4776 & 6528\\
    IT & 200 & 15k/26 & 5838 & 5938 \\
    \bottomrule
    \end{tabular}

    }
\end{table}

% models used for NER:
%     "en": "dslim/bert-base-NER-uncased",
%     "ca": "projecte-aina/roberta-base-ca-cased-ner",
%     "fr": "Babelscape/wikineural-multilingual-ner",
%     "de": "Babelscape/wikineural-multilingual-ner",
%     "es": "Babelscape/wikineural-multilingual-ner",
%     "it": "DeepMount00/Italian_NER_XXL",
%     "nl": "Babelscape/wikineural-multilingual-ner",
%     "pl": "Babelscape/wikineural-multilingual-ner",
%     "pt": "Babelscape/wikineural-multilingual-ner",

\paragraph{Biasing List Creation} \quad 
We automatically create biasing lists for target CommonVoice subsets to perform the contextual biasing experiments. For this purpose, we use BERT models from HuggingFace~\citep{wolf2020huggingface} fine-tuned on the named-entity recognition (NER) task for each language individually.\footnote{EN: \url{dslim/bert-base-NER-uncased}; DE, ES, FR: \url{Babelscape/wikineural-multilingual-ner} \citep{armengol-estape-etal-2021-multilingual_ner}; CA: \url{projecte-aina/roberta-base-ca-cased-ner}~\citep{tedeschi-etal-2021-wikineural-combined_ner2}.} The following steps are included: (1)~automatic text labeling with BERT, (2)~NEs extraction from the BERT labels, (3)~NEs lists filtering.
In Table~\ref{tab:train_and_test_sets}, one can see the statistics of NE lists per language where the size of lists with unique NEs varies from 2108 to 6949 which is rather long for contextual biasing.\footnote{The ideal size of the biasing FST is significantly influenced by the data; according to \cite{chen2019end}, performance started to decline when the number of contextual entities surpassed 1000.}
% \TODO{should we add something like "In contrast, our approach works also on bias list of large size".}
The last column of Table~\ref{tab:train_and_test_sets} shows the number of utterances per test set that contain at least one NE. This information gives an estimation of the proportion of NEs in the test sets: DE and FR sets have almost half of the utterances with NEs, at the same time, the CA set has only 17\% of those.   
%to exclude all the single-word NEs which are shorter than 5 characters

%\yn{Should we add the info about including/excluding unigrams?} \yn{can we add an equation about the inLM/OutLM? Or how do we describe this?} \yn{do we mention something about n-gram LM+Bias-list? When we merge both.}

\paragraph{Heuristics for biasing list selection} \quad Since NEs are automatically extracted with the BERT-based NER, extraction errors are inevitable. To minimize noise from potentially erroneously extracted NEs, we follow simple filtering heuristics when preparing the final biasing lists. \textbf{$H1$:} select only NEs that are composed of 1 to 4 words, and \textbf{$H2$:} remove single-word NEs shorter than 5 characters. For example, the filtering step is important to reduce such noisy outputs as short single words that often are not NEs: \textit{ich, wir, die} -- for DE, \textit{san, mar, new} -- for ES. We also tried to further filter the lists by only allowing NEs that are repeated at least twice, or NEs composed of bigram or more. However, in our experiments, only applying \textbf{$H1$} and \textbf{$H2$} was sufficient and yielded better WERs overall.
%In Table~\ref{tab:train_and_test_sets} we list the number of NEs per test set.

%- Run NER and extract only the NEs
% \begin{itemize}
%     \item select only NEs that are composed of 1 to 4 words;
%     \item remove single-word NEs shorter than \textcolor{red}{5} characters;
%     \item create two lists one with only unigrams that are repeated at least twice and one with all...
% \end{itemize}

\section{Results}
We report our results in two parts. First, we present the overall performance of models trained with PL and with different settings. The following configurations are compared: (1)~offline VS streaming TT models, (2)~ models trained on PL-only VS models with supervised data regularization, and (3)~models with different chunk sizes. Second, we report the performance of models with SF. For the \textit{baseline}, we use a Zipformer streaming model trained per each language on PL only. In addition, we also compare our results to the offline Zipformer trained on the same data and include the reference to the previous results reported on CommonVoice in \citep{whisper} (Table~\ref{tab:appendix:commonvoice-vs-whisper}).\footnote{Note that the results from \citep{whisper} are not directly comparable to ours, as we use the CV-11 version of CommonVoice and \citep{whisper} uses the version CV-7. We anyway include their results to have the previous reference point but locate them in Appendix.}

\subsection{Performance on models with PL of different quality}
\begin{figure}[t]
    \centering
    \includegraphics[width=0.99\linewidth]{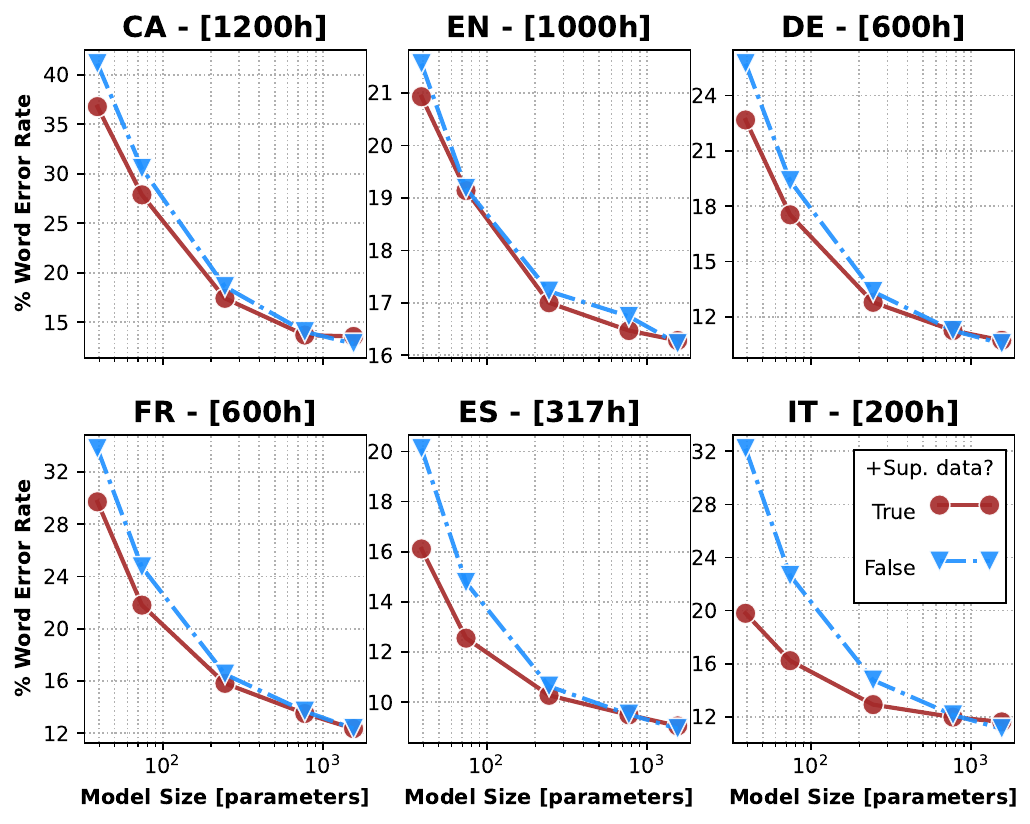}
    \caption{WERs for offline Zipformer models on six languages of CommonVoice. Models are trained with pseudo-labels from different Whisper model sizes (blue graphs). Adding 100h of supervised data during training (red graph) regularizes the training up to models with 700M params, especially for languages with less data.}
    \label{fig:baseline-results}
\end{figure}

\paragraph{Offline models} \quad In Figure~\ref{fig:baseline-results}, we present the offline results for TT models trained from scratch on PL data only in six languages (depicted by blue graphs). These models are evaluated only in a non-streaming context to determine the upper bound WERs achievable by training with PLs of varying qualities. As the size of the Whisper Model increases (shown on a log-scaled x-axis), there is a corresponding improvement in WERs, also on a log-scale. The best performance is observed for ES, with the least favourable results for EN. These results show that our approach adapts across a spectrum of PL data quantities and qualities, ranging from 200h for IT to over 1000h for CA and EN. We additionally analyzed the performance of models trained on PLs depending on how well each language is represented in the data used for training Whisper models~\citep{whisper}. Yet, no consistent effect is noticed.

\paragraph{Regularization with supervised data} \quad Red graphs in Figure~\ref{fig:baseline-results} also show WERs for offline models that along with PLs include a small amount of supervised data, up to 100h, for regularization. This strategy proves to be beneficial in cases with noisier PLs, particularly for smaller Whisper models like Whisper-tiny, Whisper-base, and Whisper-small when WER goes down for all the languages. The benefits, however, decrease or are absent with more accurate PLs generated by larger models, such as Whisper-medium and Whisper-large-v3. Thus, with our results on six languages, we can conclude that when supervised data is available, regularization is recommended for models with weak PLs and can be omitted with strong PLs. The results with 100h regularization are also available in Table~\ref{tab:appendix:commonvoice-vs-whisper} for offline models and are consistent with the performance on streaming models reported in Table~\ref{tab:commonvoice-streaming-with-supervised}.

\begin{figure*}[t]
    \centering
    \includegraphics[width=0.99\linewidth]{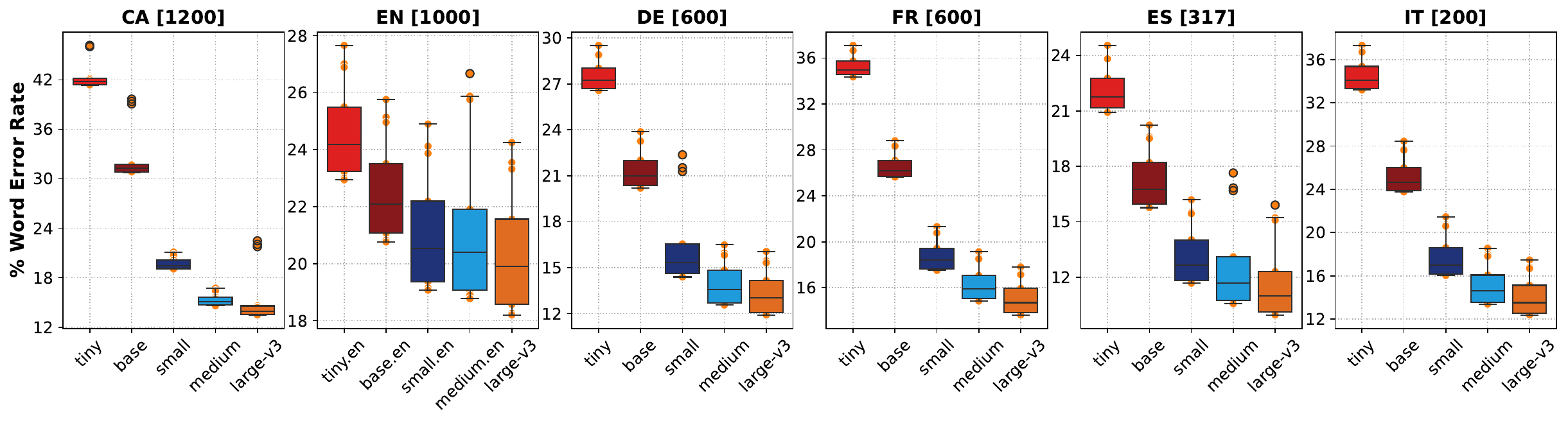}
    \caption{Box plots of WERs for six languages of CommonVoice. 
    Streaming Zipformer models are trained from scratch, with only PLs generated with different Whisper model sizes. Each box denotes 13 decoding configurations, ranging from challenging (320ms chunk with limited left context) to more relaxed (2560ms chunk with full left context) streaming settings. (Note different WER scaling on the y-axis.)}
    \label{fig:streaming-ablation}
\end{figure*}

\begin{figure}[t]
    \centering
    \includegraphics[width=0.99\linewidth]{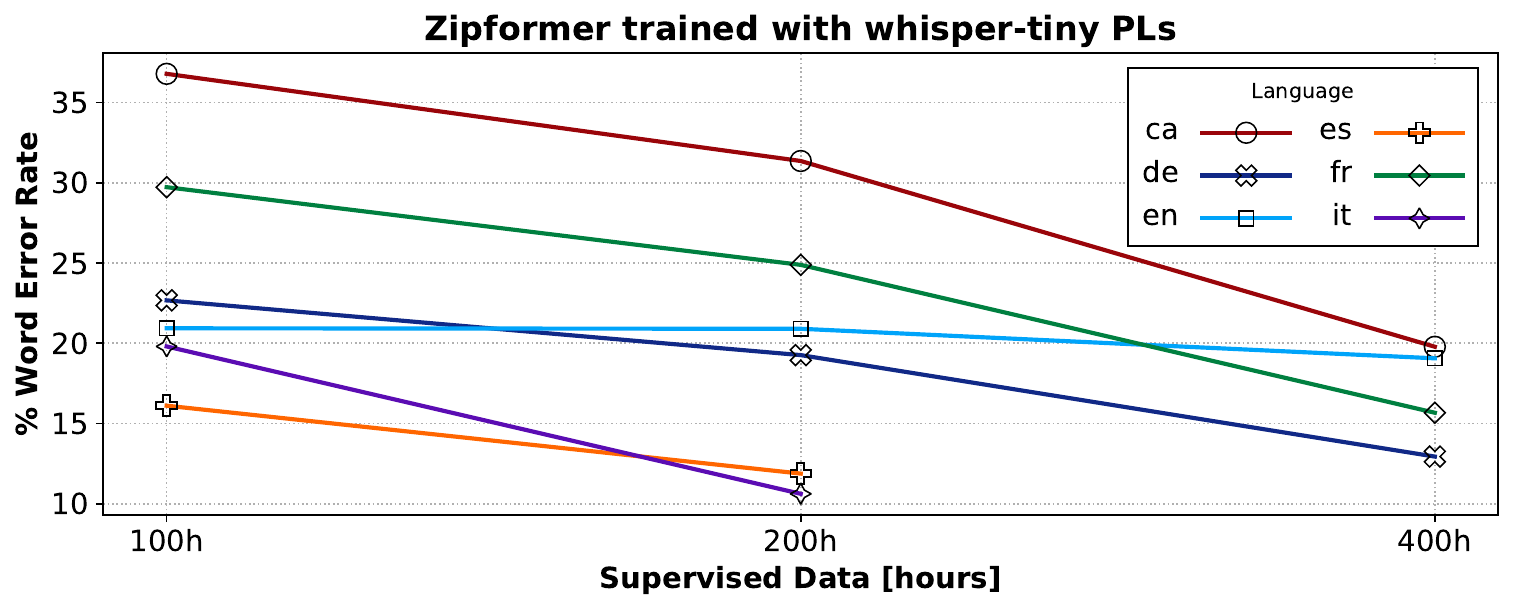}
    \caption{Ablations on WERs of Zipformer models for 6 languages of CommonVoice. We study the impact of mixing supervised data during training with pseudo-labeled of very low quality, i.e., Whisper-tiny.}
    \label{fig:dataset-size-ablation}
\end{figure}

\paragraph{Scaling-up supervised data helps on cases with very noisy PLs} \quad For the ablation experiment on mixing in more supervised data, we maintain a fixed computational budget for generating PLs and explore the extent to which supervised data can offset noisy PLs. The results are pictured in Figure~\ref{fig:dataset-size-ablation}. Using only Whisper-tiny, we train TT models from scratch for the six CommonVoice languages with over 200h of available supervised data.
%We initially base our comparison on the original WERs listed in the Whisper paper~\citep{whisper} (see Table~\ref{tab:appendix:commonvoice-vs-whisper}).
Our results show significant improvements in WER as supervised data increases from 100h to 200h and even more so up to 400h, especially in languages like Catalan, French, and Italian, which likely suffer from lower-quality PLs. For this experiment, our oracle results are from the models fully trained on the supervised data, which can be found in Table~\ref{tab:appendix:commonvoice-vs-whisper} for offline models and in Table~\ref{tab:commonvoice-streaming-with-supervised} for streaming models.
%(44\%+ WER, see Table~\ref{tab:appendix:commonvoice-vs-whisper}).

\begin{table}[]
\caption{WERs for streaming evaluation with n-gram LM and bias-lists (BL). Listed on four CommonVoice languages and two decoding configurations. The Zipformer models are trained with pseudo-labeled data from different Whisper models and 100h of supervised data (``sup.~[100h]'') from the original train set. All experiments show additive WERs improvement when adding either (or both) n-gram LM or biasing lists. 
}
\label{tab:commonvoice-streaming-with-supervised}
% \centering
\setlength{\tabcolsep}{3pt}
\resizebox{1\linewidth}{!}{

\begin{tabular}{l | cccc | cccc}
    \toprule
    & \multicolumn{4}{c|}{\textbf{cs=320ms;lf=2.5s}} & \multicolumn{4}{c}{\textbf{cs=320ms;lf=$\infty$}} \\

    \cmidrule(lr){2-5}
    \cmidrule(lr){6-9}
    \multicolumn{1}{c}{\textbf{Experiment}} & CA & DE & ES & IT & CA & DE & ES & IT \\

    \cmidrule(lr){1-9}
    \multicolumn{9}{c}{\textbf{Whisper-tiny ($39M$)}} \\
    \cmidrule(lr){1-9}
    
    Zipformer ($70M$) & 46.2 & 29.5 & 24.5 & 37.3 & 46.0 & 28.9 & 23.8 & 36.7 \\
    \quad +sup. [100h] & 38.7 & 27.0 & 21.6 & 26.7 & 38.3 & 26.4 & 21.0 & 25.9 \\
    \quad +n-gram LM & 38.6 & 26.2 & 21.2 & 25.7 & 38.2 & 25.5 & 20.5 & 25.0 \\
    \quad +bias-list & 34.4 & 26.3 & 21.3 & 25.5 & 33.9 & 25.7 & 20.5 & 24.7 \\
    \quad +n-gram LM+(BL) & 34.0 & 25.7 & 21.0 & 24.7 & 33.6 & 25.1 & 20.2 & 23.9 \\
        
    \cmidrule(lr){1-9}
    \multicolumn{9}{c}{\textbf{Whisper-base ($74M$)}} \\
    \cmidrule(lr){1-9}

    Zipformer ($70M$) & 39.7 & 23.9 & 20.2 & 28.4 & 39.4 & 23.3 & 19.5 & 27.6 \\
    \quad +sup. [100h] & 29.9 & 22.2 & 18.3 & 23.0 & 29.6 & 21.5 & 17.6 & 22.3 \\
    \quad +n-gram LM & 29.4 & 21.4 & 17.8 & 22.2 & 29.1 & 20.8 & 17.1 & 21.5 \\
    \quad +bias-list & 26.0 & 21.6 & 17.6 & 22.0 & 25.6 & 21.0 & 16.9 & 21.2 \\
    \quad +n-gram LM+(BL) & 25.7 & 21.0 & 17.2 & 21.3 & 25.2 & 20.4 & 16.5 & 20.7 \\
    
    \cmidrule(lr){1-9}
    \multicolumn{9}{c}{\textbf{Whisper-small ($244M$)}} \\
    \cmidrule(lr){1-9}
    
    Zipformer ($70M$) & 21.1 & 22.4 & 16.2 & 21.5 & 20.8 & 21.3 & 15.4 & 20.6 \\
    \quad +sup. [100h] & 20.1 & 17.8 & 16.0 & 20.1 & 19.8 & 17.2 & 15.3 & 19.3 \\
    \quad +n-gram LM & 19.6 & 17.1 & 15.5 & 19.3 & 19.3 & 16.5 & 14.8 & 18.5 \\
    \quad +bias-list & 17.7 & 17.2 & 15.4 & 19.3 & 17.3 & 16.6 & 14.7 & 18.5 \\
    \quad +n-gram LM+(BL) & 17.4 & 16.7 & 15.0 & 18.8 & 17.1 & 16.1 & 14.3 & 17.9 \\
    
    \cmidrule(lr){1-9}
    \multicolumn{9}{c}{\textbf{Whisper-medium ($769M$)}} \\
    \cmidrule(lr){1-9}

    Zipformer ($70M$) & 16.7 & 16.5 & 17.6 & 18.6 & 16.4 & 15.8 & 16.7 & 17.8 \\
    \quad +sup. [100h] & 16.5 & 16.6 & 14.9 & 19.4 & 16.2 & 15.8 & 14.3 & 18.4 \\
    \quad +n-gram LM & 16.1 & 15.8 & 14.6 & 18.6 & 15.8 & 15.1 & 13.9 & 17.7 \\
    \quad +bias-list & 14.8 & 16.0 & 14.6 & 18.6 & 14.5 & 15.3 & 13.9 & 17.7 \\
    \quad +n-gram LM+(BL) & 14.6 & 15.5 & 14.3 & 18.1 & 14.3 & 14.8 & 13.6 & 17.2 \\
    
    \cmidrule(lr){1-9}
    \multicolumn{9}{c}{\textbf{Whisper-large-v3 ($1.5B)$}} \\
    \cmidrule(lr){1-9}

    Zipformer ($70M$) & 22.5 & 16.1 & 15.9 & 17.5 & 21.8 & 15.3 & 15.1 & 16.7 \\
    \quad +sup. [100h] & 16.6 & 16.3 & 14.6 & 18.4 & 16.4 & 15.6 & 14.0 & 17.6 \\
    \quad +n-gram LM & 16.2 & 15.6 & 14.2 & 17.6 & 16.0 & 14.9 & 13.6 & 16.8 \\
    \quad +bias-list & 15.0 & 15.8 & 14.1 & 17.8 & 14.8 & 15.1 & 13.5 & 17.0 \\
    \quad +n-gram LM+(BL) & 14.8 & 15.3 & 13.8 & 17.2 & 14.5 & 14.6 & 13.2 & 16.4 \\

    \cmidrule(lr){1-9}
    \multicolumn{9}{c}{\textbf{Baseline streaming Zipformer (only supervised data)}} \\
    \cmidrule(lr){1-9}
    Zipformer ($70M$) & 7.8 & 13.8 & 13.5 & 17.5 & 7.6 & 13.1 &12.8 & 16.6 \\
    
    \bottomrule
\end{tabular}
}
\end{table}

\begin{table}[]
\caption{WERs for six CommonVoice languages. The Zipformer offline models are trained with pseudo-labeled data from different Whisper models. We also report WERs when a small amount of supervised data is added during training, denoted as \texttt{``sup.~[100h]''}. Note that the transducer models are trained from scratch in $\sim$1 day GPU time.
%\TODO{ Whisper results are in CV-7, but ours in CV-11.} 
}
\label{tab:appendix:commonvoice-vs-whisper}
% \centering
\setlength{\tabcolsep}{3pt}
\resizebox{1\linewidth}{!}{

\begin{tabular}{l | cccccc}
    \toprule
    & \multicolumn{6}{c}{\textbf{Language [hours]}} \\
    \cmidrule(lr){2-7}
    \multicolumn{1}{c|}{\textbf{\textbf{Experiment}}} & CA & EN & DE & FR & ES & IT \\
    % \midrule
    & 1200 & 1000 & 600 & 600 & 317 & 200 \\
    
    \cmidrule(lr){1-7}
    \multicolumn{7}{c}{\textbf{Whisper-tiny ($39M$)}} \\
    \cmidrule(lr){1-7}
    \citet{whisper} & 51.0 & 28.8 & 34.5 &  49.7 & 30.3 & 44.5 \\
    Zipformer ($70M$) & 41.1 & 21.5 & 25.7 & 33.8 & 20.1 & 32.2 \\ 
    \quad +sup. [100h] & 36.8 & 20.9 & 22.6 & 29.7 & 16.1 & 19.8 \\
    \quad +n-gram LM & 32.4 & 21.0 & 22.1 & 31.3 & 15.9 & 18.7 \\
    \quad +n-gram LM+(BL)  & 32.0 & 20.7 & 21.5 & 30.8 & 15.6 & 18.0 \\

    \midrule
    \multicolumn{7}{c}{\textbf{Whisper-base ($74M$)}} \\
    \cmidrule(lr){1-7}
    \citet{whisper} & 39.9 & 21.9 &  24.5 & 37.3 & 19.6  & 30.5\\
    Zipformer ($70M$) & 30.5 & 19.2 & 19.4 & 24.7 & 14.8 & 22.7 \\
    \quad +sup. [100h] & 27.9 & 19.1 & 17.5 & 21.8 & 12.6 & 16.3 \\
    \quad +n-gram LM & 24.0 & 19.1 & 17.0 & 22.2 & 12.2 & 15.5 \\
    \quad +n-gram LM+(BL)  & 23.7 & 18.8 & 16.4 & 21.7 & 11.8 & 14.8 \\

    \midrule
    \multicolumn{7}{c}{\textbf{Whisper-small ($244M$)}} \\
    \cmidrule(lr){1-7}    
    \citet{whisper} & 23.8 & 14.5 & 13.0 & 22.7&  10.3& 16.0\\
    Zipformer ($70M$) & 18.6 & 17.1 & 13.4 & 16.5 & 10.7 & 14.8 \\
    \quad +sup. [100h] & 17.4 & 16.9 & 12.8 & 15.8 & 10.2 & 12.9 \\
    \quad +n-gram LM & 15.1 & 16.7 & 12.4 & 15.8 & 10.0 & 12.4 \\
    \quad +n-gram LM+(BL) & 14.9 & 16.4 & 11.9 & 15.4 & 9.7 & 11.8 \\

    \midrule
    \multicolumn{7}{c}{\textbf{Whisper-medium ($769M$)}} \\
    \cmidrule(lr){1-7}
    \citet{whisper} & 16.4 & 11.2 &  8.5 &  16.0 &  6.9& 9.4\\
    Zipformer ($70M$) & 14.0 & 16.7 & 11.3 & 13.7 & 9.5 & 12.1 \\
    \quad +sup. [100h] & 13.7 & 16.4 & 11.3 & 13.5 & 9.5 & 12.0 \\
    \quad +n-gram LM & 12.1 & 16.2 & 10.9 & 13.2 & 9.3 & 11.5 \\
    \quad +n-gram LM+(BL)  & 11.9 & 15.9 & 10.4 & 12.9 & 9.0 & 11.1 \\

    \midrule
    \multicolumn{7}{c}{\textbf{Whisper-large-v3 ($1.5B)$}} \\
    \cmidrule(lr){1-7}
    \citet{whisper} & 14.1 & 9.4 &  6.4 & 13.9 &  5.6& 7.1\\
    Zipformer ($70M$) & 12.8 & 16.2 & 10.5 & 12.4 & 8.9 & 11.1 \\
    \quad +sup. [100h] & 13.6 & 16.3 & 10.7 & 12.4 & 9.0 & 11.6 \\
    \quad +n-gram LM & 12.1 & 16.0 & 10.4 & 12.0 & 8.9 & 11.3 \\
    \quad +n-gram LM+(BL)  & 11.8 & 15.6 & 10.0 & 11.6 & 8.6 & 10.8 \\

    \midrule
    \multicolumn{7}{c}{\textbf{Baseline offline Zipformer (only supervised data)}} \\
    \cmidrule(lr){1-7}
    Zipformer ($70M$) & 4.9 & 14.5 & 8.5 & 10.7 & 8.1 & 10.2 \\

    \bottomrule
\end{tabular}
}
\end{table}

\paragraph{Low-latency streaming decoding} \quad Figure~\ref{fig:streaming-ablation} lists the streaming decoding results across six CommonVoice languages, testing 13 different decoding configurations (see~\ref{subsec:tt_training}). We establish an upper performance bound with models tested in non-streaming mode and also include a box plot for each TT model trained with PLs derived from various Whisper model sizes. 
%Larger Whisper models consistently yield lower mean WERs across all configurations, 
The results show how model performance can fluctuate under different streaming conditions, with smaller chunk sizes or limited left context posing greater challenges.

The results with the configuration with cs=320ms and lf=2.5s are also reported in Tables~\ref{tab:commonvoice-streaming-with-supervised} and \ref{tab:appendix:commonvoice-streaming-only-pl} and demonstrate consistently better performance when the full left context is used. This tendency stays independent of language and SF.

%\TODO{response to your comment above: Oh yeah, we need to split the results in this way: the figure only covers the models trained with PL and those are the baseline results. However, the table covers the results on the streaming setting when we add n-gram LM and biasing list. The table is just extra numbers that should say something like "hey, we also tested the biasing and n-gram approaches if you're doing streaming decoding. We only selected 4 languages and tested 2 different streaming very challenging settings. Maybe we need to make two subsections or paragraphs for this.}

\subsection{SF with n-gram LM brings substantial WER reductions on challenging scenarios - and decoding analysis}

Performance on different models and languages with SF is presented in Table~\ref{tab:commonvoice-streaming-with-supervised}. Zipformer models in the table are our baselines, i.e., streaming models trained on PL data. We use the further improved models with an additional 100h of supervised data for decoding with SF. For all the languages, we can see a WER decrease when decoded with an external LM. The WER also always improves when context biasing with NEs is introduced. It is an important observation since all NEs are extracted automatically with no human supervision involved. Moreover, all biasing lists are rather long, which often is an obstacle to improvement when biasing methods are used~\citep{chen2019end}. Nevertheless, our approach proves to work on biasing lists of large sizes as well. According to our results, external LM and context NEs fusions are complementary methods, gaining the best WER when they are combined during decoding.

The improvement with SF is consistent through the languages and has the biggest impact when models are trained on weaker PLs generated from the smaller Whisper models. This behavior is expected, as the models that saw less training data have more potential to still benefit from any additional data given during regularization and/or decoding. On the other hand, the improvement decreases with the PLs generated by Whisper-medium and Whisper-large-v3.

Another remarkable observation is that the models trained on PLs are more competitive with the models trained on the supervised data only when less training data is given. For example, CA language has 1200h of training data and supervised models are considerably winning over the PL models even after all the improvements we introduce: 7.8\% VS 14.8\% for supervised and PL models correspondingly.\footnote{Here and below in this paragraph, we report WERs for the configuration with cs=320ms and lf=2.5s. We observe the same tendency in the results with the other configuration as well.} When double less training data is used for DE language, i.e., 600h, the difference is less prominent but still considerable: 13.8\% VS 15.3\% for supervised and PL models correspondingly. When the amount of training data is further reduced to 317h for ES and 200h for IT, we observe either little or no degradation from supervised models to PL models: 13.5\% VS 13.8\% for ES and 17.5\% VS 17.2\% for IT for supervised and PL models correspondingly. These results illustrate well the advantages and strengths of the proposed framework and methods for the low-resource scenarios. Due to space constraints, in Table~\ref{tab:commonvoice-streaming-with-supervised}, we show the performance only on four languages; SF impact on offline models for all six languages can be found in Table~\ref{tab:appendix:commonvoice-vs-whisper}. 

% \subsection{Student final performance scales in a similar proportion as model size} - insert plot about how model scale affect final WER
% \paragraph{Extreme Low-resource scenario} In this case, we only have few hours of pseudo-labeled data and only out-of-domain text corpus for shallow fusion, we use MLS text to train n-gram LMs

\section{Conclusions}

In this work, we propose a framework to meet the challenge of training streaming ASR systems with few-to-none supervised data by leveraging PLs from foundational speech models.
We conduct a thorough examination of the efficacy of PL-based TT models across various dimensions, including offline and chunk-wise decoding for streaming applications, and the influence of FSM size on the TT model's WERs. We introduce robust heuristics to filter out unreliable and hallucinated PLs. Our findings reveal that TT models can be effectively trained from scratch on noisy PLs. We managed to further improve the performance of models trained with weak pseudo-labels (generated by Whisper-tiny, -base, and -small) by adding regularization with different amounts of supervised data. Additionally, we prove that decoding with the shallow fusion of external n-gram LM and automatically generated named entities always improves the performance of models, independent of the quality of pseudo-labels.
% \TODO{to update and add other results}

% \clearpage

\section*{Limitations}

One of the limitations of the paper is that the data from the CommonVoice dataset is read speech that can considerably differ from spontaneous speech and unprepared conversations. Our choice was mostly due to the possibility of testing our framework on six different languages and in this regard, CommonVoice suited us well. Besides this, models for each language were trained on a different amount of data (from 200h to 1200h) that demonstrated different impacts of the proposed methods. However, no experiments were done to see the performance with different amounts of train data within each language.

Another limitation of the paper is that despite focusing mostly on the streaming ASR models, we provide no results on the execution time. This information would be especially important for the shallow fusion experiments. Although we noted good time performance of the proposed shallow fusion implementation for offline models, the evaluation for streaming models is missing.

\section*{Ethical Considerations}
All speech data sets we use have anonymous speakers. We do not have any access to nor try to create any PII of speakers.

\section*{Acknowledgements}
This work was supported by the Idiap~\&~Uniphore collaboration project.
Part of the work was also support supported by EU Horizon 2020 project ELOQUENCE\footnote{\url{https://eloquenceai.eu/}} (grant number 101070558).

\bibliography{bib_new}

\newpage
\appendix
\section{Streaming decoding configurations}
\label{sec:appendix:streaming-configurations}

We perform a swipe of streaming decoding evaluations under multiple low-latency settings. We evaluate the following configurations:

\begin{itemize}
    \item Decode chunk size = 320ms with left context of 2560ms, 5120ms and full;
    \item decode chunk size = 640ms with left context of 2560ms, 5120ms and full;
    \item decode chunk size = 1280ms with left context of 2560ms, 5120ms and full;
    \item decode chunk size = 2560ms with left context of 2560ms, 5120ms and full.
\end{itemize}

\noindent The overall results are reported in Figure~\ref{fig:streaming-ablation} for each of the proposed languages.

\section{Extended results for models trained on PL data}
\label{sec:appendix:extended-results}

\begin{table}[]
\caption{WERs for streaming evaluation with n-gram LM and bias-lists (BL). Listed on four CommonVoice languages and two decoding configurations. The Zipformer models are trained with only pseudo-labeled data from different Whisper models. All experiments show additive WERs improvement when adding either (or both) n-gram LM or biasing lists.
%\TODO{there is some weird behaviour in the CA model trained with Whisper-large-v3 and evaluated on streaming setting. We can maybe leave this table in the appendix and only use it in the main paper the one that uses 100h of regularization}
}
\label{tab:appendix:commonvoice-streaming-only-pl}
\centering
\setlength{\tabcolsep}{3pt}
\resizebox{1\linewidth}{!}{

\begin{tabular}{l | cccc | cccc}
    \toprule
    & \multicolumn{4}{c|}{\textbf{cs=320ms;lf=2.5s}} & \multicolumn{4}{c}{\textbf{cs=320ms;lf=$\infty$}} \\

    \cmidrule(lr){2-5}
    \cmidrule(lr){6-9}
    \multicolumn{1}{c}{\textbf{Experiment}} & CA & DE & ES & IT & CA & DE & ES & IT \\

    \cmidrule(lr){1-9}
    \multicolumn{9}{c}{\textbf{Whisper-tiny ($39M$)}} \\
    \cmidrule(lr){1-9}
    
    Zipformer ($70M$) & 46.2 & 29.5 & 24.5 & 37.3 & 46.0 & 28.9 & 23.8 & 36.7 \\
    \quad +n-gram LM & 46.0 & 29.0 & 24.0 & 36.5 & 45.8 & 28.4 & 23.2 & 35.9 \\
    \quad +bias-list & 43.7 & 28.9 & 23.7 & 36.0 & 43.6 & 28.4 & 23.0 & 35.3 \\
    \quad +n-gram LM+(BL) & 43.6 & 28.4 & 23.3 & 35.3 & 43.5 & 28.0 & 22.6 & 34.6 \\  
    
    \cmidrule(lr){1-9}
    \multicolumn{9}{c}{\textbf{Whisper-base ($74M$)}} \\
    \cmidrule(lr){1-9}

    Zipformer ($70M$) & 39.7 & 23.9 & 20.2 & 28.4 & 39.4 & 23.3 & 19.5 & 27.6 \\
    \quad +n-gram LM & 39.1 & 23.2 & 19.8 & 27.5 & 38.7 & 22.5 & 19.0 & 26.7 \\
    \quad +bias-list & 36.2 & 23.3 & 19.6 & 27.2 & 36.0 & 22.7 & 18.9 & 26.4 \\
    \quad +n-gram LM+(BL) & 36.0 & 22.7 & 19.3 & 26.5 & 35.7 & 22.2 & 18.5 & 25.7 \\    
    \cmidrule(lr){1-9}
    \multicolumn{9}{c}{\textbf{Whisper-small ($244M$)}} \\
    \cmidrule(lr){1-9}

    Zipformer ($70M$) & 21.1 & 22.4 & 16.2 & 21.5 & 20.8 & 21.3 & 15.4 & 20.6 \\
    \quad +n-gram LM & 20.8 & 21.8 & 15.8 & 20.7 & 20.5 & 20.8 & 15.0 & 19.8 \\
    \quad +bias-list & 20.0 & 21.7 & 15.6 & 20.5 & 19.7 & 20.6 & 14.8 & 19.7 \\
    \quad +n-gram LM+(BL) & 19.8 & 21.3 & 15.2 & 20.0 & 19.6 & 20.2 & 14.5 & 19.1 \\ 
    
    \cmidrule(lr){1-9}
    \multicolumn{9}{c}{\textbf{Whisper-medium ($769M$)}} \\
    \cmidrule(lr){1-9}

    Zipformer ($70M$) & 16.7 & 16.5 & 17.6 & 18.6 & 16.4 & 15.8 & 16.7 & 17.8 \\
    \quad +n-gram LM & 16.4 & 15.8 & 17.4 & 17.8 & 16.1 & 15.1 & 16.4 & 17.0 \\
    \quad +bias-list & 16.1 & 16.0 & 17.0 & 17.8 & 15.9 & 15.3 & 16.0 & 17.1 \\
    \quad +n-gram LM+(BL) & 15.9 & 15.6 & 16.8 & 17.3 & 15.7 & 14.9 & 15.8 & 16.5 \\

    \cmidrule(lr){1-9}
    \multicolumn{7}{c}{\textbf{Whisper-large-v3 ($1.5B)$}} \\
    \cmidrule(lr){1-9}

    Zipformer ($70M$) & 22.5 & 16.1 & 15.9 & 17.5 & 21.8 & 15.3 & 15.1 & 16.7 \\
    \quad +n-gram LM & 22.4 & 15.4 & 15.6 & 16.8 & 21.7 & 14.7 & 14.9 & 16.0 \\
    \quad +bias-list & 21.9 & 15.6 & 15.5 & 16.9 & 21.1 & 14.9 & 14.8 & 16.1 \\
    \quad +n-gram LM+(BL) & 22.1 & 15.2 & 15.3 & 16.4 & 21.3 & 14.5 & 14.6 & 15.6 \\

    \cmidrule(lr){1-9}
    \multicolumn{9}{c}{\textbf{Baseline streaming Zipformer (only supervised data)}} \\
    \cmidrule(lr){1-9}
    Zipformer ($70M$) & 7.8 & 13.8 & 13.5 & 17.5 & 7.6 & 13.1 &12.8 & 16.6 \\
    \bottomrule
\end{tabular}
}
\end{table}

\paragraph{Offline models evaluation} \quad Table~\ref{tab:appendix:commonvoice-vs-whisper} shows the WERs for Zipformer offline models trained on six CommonVoice languages with either solely PLs or a mix of PLs and a small amount of supervised data (100h). These extended results correspond to those depicted in Figure~\ref{fig:baseline-results} in the main paper.

% Previos work as in~\citet{ma2024embarrassingly}, have shown very close performance (WERs) between Whisper-medium and Whisper-large. In some cases, the latter yielding workse performnace. We

\paragraph{Streaming models evaluation} \quad Table~\ref{tab:appendix:commonvoice-streaming-only-pl} shows the WERs for Zipformer streaming models trained on four CommonVoice languages with solely PLs and evaluated on two different streaming configurations.

\section{Filtering stage}
\label{sec:appendix-filtering-stage}

As part of our efforts to reduce the amount of the hallucinated or low-quality pseudo-labels, we propose to filter out data based on some heuristics, as described in Section~\ref{sec:experimental-setup}.

In Table~\ref{tab:appendix:max-word-length-pl} we list the exact statistics of the maximum number of characters allowed per pseudo-labeled word for each dataset from CommonVoice. Note that languages that join words, such as German (DE) have a substantially larger threshold. Note that if a single word of the full pseudo-labeled utterance meets the threshold, we discard the entire sample.

\begin{table}[h]
\caption[Maximum number of characters allowed in each pseudo-labeled word with Whisper.]{Maximum number of characters allowed in each pseudo-labeled word with Whisper. 
}
\label{tab:appendix:max-word-length-pl}
\centering
\begin{tabular}{cccccc}
    \toprule
    \textbf{CA} & \textbf{EN} & \textbf{DE} & \textbf{FR} & \textbf{ES} & \textbf{IT} \\
    \midrule
    16 & 16 & 30 & 20 & 25 & 22 \\
    \bottomrule
\end{tabular}
\end{table}

% \clearpage

\section{Call-center speech use case}
\label{sec:appendix-call-center-use-case}

We also evaluate our approach on a particularly important use case for industrial applications. Here, we are given 1.7k hr of unlabeled audio and our task is to train a TT system from scratch without incurring costly labeling for supervised training. This is a challenging scenario because Whisper models might not perform as well as in benchmarks due to, unseen noise or artifacts, accent or simply due to domain shift. Below, we define the database and the steps followed. 

\subsection{Call-center database} 

We employ a collection of unlabeled two-channel agent-user conversations of more than 10 minutes long from the call center domain. In total, there are 12.8k WAV audio files, i.e., $\sim$1728\,hr. We use a 54-min test set with gold annotations to evaluate our system. We generate pseudo-labels with WhisperX pipeline (\S\ref{subsec:pseudo-labeling-with-whisper}), though we slightly modify the VAD step to only allow up to 5 seconds of contiguous silence between contiguous segments. This process leads to 735 hrs of pure pseudo-labeled audio. 

\subsection{Baseline performance}

\begin{figure}[t]
    \centering
    \includegraphics[width=0.99\linewidth]{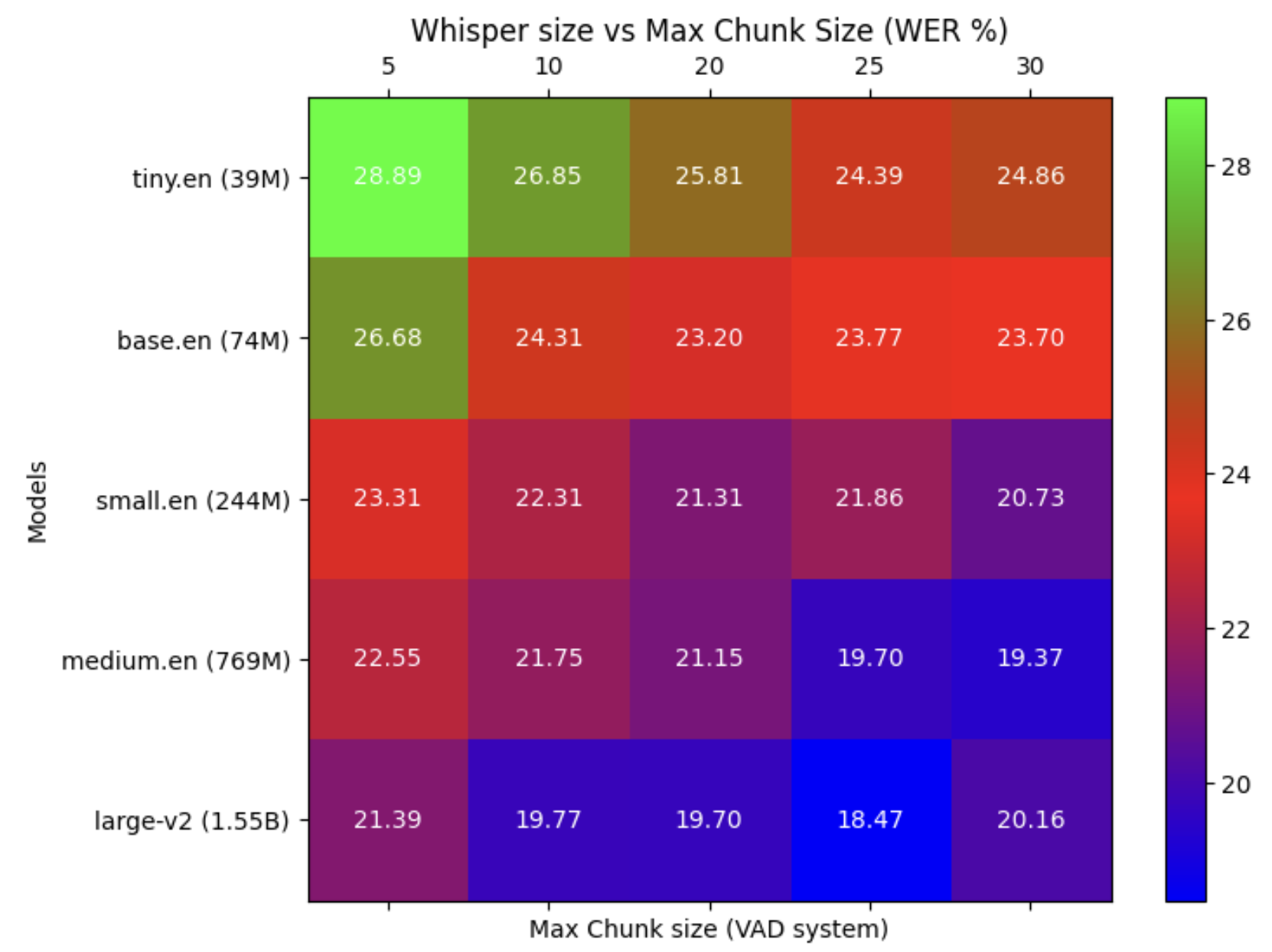}
    \caption{WERs on the test set with different Whisper model configurations and chunk sizes of the VAD model.}
    \label{fig:whisper-vs-chunk-size}
\end{figure}

In Figure~\ref{fig:whisper-vs-chunk-size} we list a matrix with the WERs obtained by varying the Whisper model size and the maximum chunk size for the cut and merge step from WhisperX~\cite{bain23_whisperX}. See more information in \S\ref{subsec:pseudo-labeling-with-whisper}. Increasing model size yields better WERs while having 25-second long segments produces lower WERs overall. This is expected as the Whisper model is trained with audios of $\sim$30 seconds long~\cite{whisper}.

\subsection{Filtering Stage}

We perform an exhaustive filtering stage to remove potential low-quality data. This step further reduces the dataset to 510 hours, i.e., a 30\% relative reduction. We use similar heuristics as in \S\ref{subsec:pseudo-labeling-with-whisper} to reduce the hallucinated hypotheses.\footnote{An example of a hallucinated hypothesis: \texttt{utt-id-01 \textit{let me try to turn my flashlight on okay w b a d w b a d w w w w}}.} 

\subsection{Additional supervised data}

We use GigaSpeech~\cite{chen21o_gigaspeech_corpus} (GS) L subset (2.5k hours), full LibriSpeech~\cite{panayotov_librispeech_corpus} (LS) train set and CommonVoice English~\cite{ardila2019_commonvoice_corpus} (CV) train subset (1.5k hours) as extra datasets during training. This aims to regularize the training phase. In total, we use 5k hours of speech as extra datasets, while 510 hours are set as the target domain set.

\subsection{Pseudo-labeled data filtering}

As we aim to develop an ASR system as fast as possible, we developed a process to select a subset of the PL database smartly. 
We extract acoustic and text-based metadata from each $\{X,Y^*\}$ pair. The acoustic metrics (1)~STOI, PESQ and SI-SDR are computed with TorchAaudio-SQUIM~\cite{torchaudio_squim}; (2)~perplexity is computed with GPT2~\cite{radford2019_gpt2} using HuggingFace~\cite{wolf2020huggingface,lhoest2021hf_datasets} and (3)~a pseudo-edit-distance metric computed by comparing different Whisper model outputs, i.e., WER metric: \texttt{whisper-tiny}:\textbf{hypothesis} \& \texttt{whisper-large-v2}:\textbf{reference}. 

\subsection{Experiments}

We perform two experiments for the call center use case. First, in the baseline scenario, we filter out (or select) a subset from the original PL dataset by using one or multiple metrics, e.g., perplexity and SI-SDR threshold. We use the remaining dataset for ASR training. Second, we are presented with a fixed computational budget that limits the final dataset size for model training. This leads us to select a smaller portion of the PL dataset based on i) random selection or ii) sorting the PL dataset by one metric (e.g., SI-DR) and then selecting the top samples that meet the allowed computational budget.

\section{Data Selection Based on Metrics}
\label{sec:appendix:data-selection}

This is the baseline scenario, where we filter the PL dataset by one or multiple metrics, and then we use the remaining dataset for ASR training. The results of this approach are listed in Table~\ref{tab:results-based-on-metrics}. Experiment 0) shows the WERs when using all the PL dataset, which serves as the baseline. From Exp 1) to 5) we run several filtering strategies, with some proposed metrics. Furthermore, we note that Exp 3) shows the best WERs while using 25\% less data than Exp 0). This translates to faster training and convergence of the Zipformer model. In conclusion, these early experiments indicate that better WERs can be attained with fewer data points when a smart policy is in place. For instance, 0.5\% absolute WER reduction, i.e., 13.9\% WER (Exp 0) $\rightarrow$ 13.4\% WER (Exp 3) from Table~\ref{tab:results-based-on-metrics}.

\begin{table}[h]
\caption{WERs for Zipformer models trained for 20 epochs with different data selection policies. Note that all experiments use the multi-dataset training recipe unless otherwise specified. $^{\dagger}$Metric computed from comparing hypothesis between Whisper \textit{tiny} and \textit{large-v2}.
}
\label{tab:results-based-on-metrics}

\centering
\setlength\tabcolsep{2pt} % default value: 6pt
\resizebox{1\linewidth}{!}{
\begin{tabular}{ l | ccccc | cc }
\toprule
    \textbf{Exp} & \multicolumn{5}{c|}{\textbf{Data selection policy}} & \textbf{Dataset} & \textbf{WER} \\
    \cline{2-6}
    & PPL & STOI & SI-SDR & WER$^{\dagger}$ & BLEU & \textbf{Size} & \\
    \midrule 
    -) & \multicolumn{5}{l|}{ALL data (no additional data)} & 510 & 15.1 \\
    0) & \multicolumn{5}{l|}{ALL data (baseline model)} & 510 & 13.9 \\
    \midrule
    1) & $\leq$ 500 & $\leq$ 0.7 & $\geq$ 15 & - & - & 210 & 15.1 \\
    2) & $\leq$ 800 & $\leq$ 0.3 & $\geq$ 5 & - & - &437 & 13.5 \\

    3) & - & - & - & $\leq$ 25\% & - & 387 & \textbf{13.4} \\
    4) & - & - & - & - & $\geq$ 50 & 428 & 13.9 \\

\bottomrule
\end{tabular}
}
\end{table}

\paragraph{Fixed Computational Budget} \quad In this setting, we are presented with a fixed computational budget, i.e., limited by the dataset sized for model training. This leads to selecting a smaller portion of the PL dataset based on i) random selection or ii) sorting the PL dataset by one metric (e.g., SI-DR) and then selecting the top samples meeting the allowed budget.  
These results are listed in Table~\ref{tab:results-based-on-datasize}. We can see significant WERs improvements up to the 200h of training. After this point, bringing more PL data at training time does not improve significantly WERs. In addition, we can conclude that none of the proposed sorting metrics is significantly better than random selection for ASR training when a fixed computational budget is imposed. \textbf{There are several hypotheses that can justify these results, as follows:}
\begin{enumerate}[nosep,topsep=-\parskip]
    \item The amount of PL data brings more WERs reductions than the proposed sorting metrics;
    \item pseudo-labels from Whisper large-v2 are of sufficiently good quality, close to gold annotation levels;
    \item the filtering stage is already removing most of the noisy and/or hallucinated PLs, i.e., the remaining 510-hour subset is already of good quality overall; 
    \item the proposed sorting metrics are not sufficiently discriminative for selecting the data required for the downstream application, i.e., random selection leads to lower WERs in some cases;
    \item using supervised data at training time brings important regularization, thus minimizing the issue of using noisy PLs.
\end{enumerate}

\begin{table}[t]
\caption{WERs for Zipformer models trained for 10 epochs with different computational budgets w.r.t amount of data. $^{\dagger}$delta of relative WER reduction 50h $\rightarrow$ 400h.}
\label{tab:results-based-on-datasize}

\centering
\setlength\tabcolsep{3pt} % default value: 6pt
\resizebox{1\linewidth}{!}{
\begin{tabular}{ l | l | ccccc | c }
    \toprule
    & \multicolumn{1}{c|}{\textbf{Sorting}} & \multicolumn{5}{c|}{\textbf{Dataset size}} & $\Delta^{\dagger}$\\
    \cmidrule{3-7}
    & \multicolumn{1}{c|}{\textbf{Metric}} & 50h & 100h & 200h & 300h & 400h &  \\
    \midrule 

    0) & ALL data (baseline) & \multicolumn{5}{c|}{[510h] 13.9\% WER} & \\
    \midrule
    1) & WER ($\downarrow$) & 30.7 & \textbf{19.3} & 15.3 & 15.0 & \textbf{13.8} & 55\%  \\
    2) & Perplexity ($\downarrow$) & 31.3 & 21.3 & 17.2 & 14.6 & 14.0 & 55\% \\
    3) & STOI ($\uparrow$) & 33.0 & 21.7 & 16.6 & 15.9 & 13.9 & 57\% \\
    4) & Random selection & \textbf{30.0} & 19.7 & \textbf{14.9} & \textbf{14.2} & 13.9 & 53\% \\
    \bottomrule
\end{tabular}
}
\end{table}

\noindent \textbf{The filtering stage is key for model training.} \quad We confirmed this hypothesis by training a Zipformer model with multi-dataset training on the unfiltered PL dataset, i.e., a 735-hour subset. To our surprise, the model performance, even though seeing more data than Exp 0 (Table~\ref{tab:results-based-on-datasize} and Table~\ref{tab:results-based-on-metrics}), did not reach acceptable WERs, e.g., 30\%+ WER. Further research down this line will shed light on what are the best practices for selecting representative data for training, including filtering of hallucinated PLs. Note that selecting or sorting PLs might be of less importance as the dataset size increases.

\end{document}